\documentclass{bmvc2k}

\usepackage{graphicx}
\usepackage{multirow}
\usepackage{color}
\usepackage{amssymb,amsmath,amsthm,enumerate,threeparttable,bm,subfigure,graphicx}
\usepackage{amsmath, amsfonts}
\usepackage{algorithm}
\usepackage{algorithmic}
\usepackage{marvosym}

\title{Alleviating Noisy-label Effects in Image Classification via Probability Transition Matrix}

\addauthor{Ziqi Zhang}{zq-zhang18@mails.tsinghua.edu.cn}{1}
\addauthor{Yuexiang Li \textsuperscript{\Letter} }{vicyxli@tencent.com}{2}
\addauthor{Hongxin Wei}{hongxin001@e.ntu.edu.sg}{3}
\addauthor{Kai Ma}{kylekma@tencent.com}{2}
\addauthor{Tao Xu \textsuperscript{\Letter} }{taoxu@tsinghua.edu.cn}{1}
\addauthor{Yefeng Zheng}{ yefengzheng@tencent.com}{2}

\addinstitution{
Tsinghua-Berkeley Shenzhen Institute, Tsinghua University\\
Shenzhen, China
}
\addinstitution{
Tencent Jarvis Lab\\
Shenzhen, China
}
\addinstitution{
School of Computer Science and Engineering, Nanyang Technological University\\
Singapore
}

\runninghead{ZHANG, LI ET AL.}{Alleviate Noisy-label via Probability Transition Matrix}

\begin{document}

\maketitle

\begin{abstract}

Deep-learning-based image classification frameworks often suffer from the noisy label problem caused by the inter-observer variation. 
Recent studies employed learning-to-learn paradigms (\emph{e.g.,} Co-teaching and JoCoR) to filter the samples with noisy labels from the training set. However, most of them use a simple cross-entropy loss as the criterion for noisy label identification. The hard samples, which are beneficial for classifier learning, are often mistakenly treated as noises in such a setting, since both the hard samples and the ones with noisy labels lead to a relatively larger loss value than the easy cases. In this paper, we propose a plugin module, namely noise ignoring block (NIB), consisting of a probability transition matrix and an inter-class correlation (IC) loss, to separate the hard samples from the mislabeled ones, and further boost the accuracy of image classification network trained with noisy labels. Concretely, our IC loss is calculated as Kullback-Leibler divergence between the network prediction and the accumulative soft label generated by the probability transition matrix.
Such that, with lower value of IC loss, the hard cases can be easily distinguished from mislabeled cases. Extensive experiments are conducted on natural and medical image datasets (CIFAR-10 and ISIC 2019). The experimental results show that our NIB module consistently improves the performances of the state-of-the-art robust training methods.
\end{abstract}

\section{Introduction}
\label{sec:intro}
Witnessing the success of deep neural networks (DNNs) for computer vision tasks \cite{masi2018deep,tan2020efficientdet}, an increasing number of researchers began to implement deep-learning-based approaches for image classification tasks.
However, due to the inter-observer variation, the training set often contains noisy labels, which may significantly degrade the model performance, \emph{i.e.,} the memorization effects of DNNs \cite{zhang2019understanding}.
Recent studies have proposed algorithms to train deep learning models with noisy labels, which can be mainly grouped into two categories, \emph{i.e.,} noise estimation \cite{liu2015classification,menon2015learning,sakai2017semi} and instance selection \cite{jiang2018mentornet,malach2017decoupling,han2018co,wei2020combating}. The former approaches detect the samples with noisy labels and accordingly revise their labels via the noise transition matrix or noise rate derived from the prior-knowledge, which limits their application only to the data with known noise characteristics. The latter categoty addresses the problem by proposing learning-to-learn selection paradigms (\emph{e.g.,} MentorNet \cite{jiang2018mentornet}, Co-teaching \cite{han2018co}, and JoCoR \cite{wei2020combating}) for the robust learning with unknown noisy labels. Generally, they treat the samples with the smaller loss as `clean', and exploit them for network training.
The existing learning-to-learn approaches share a common drawback---most of them use a simple cross-entropy loss as the criterion for noisy label identification. The hard samples, which are beneficial for classifier learning, are often mistakenly treated as noises in such a setting, since both the hard samples and the ones with noisy labels lead to a relatively larger loss value than the easy cases. 

In this paper, we propose a simple-yet-effective plugin module, namely noise ignoring block (NIB), which can be easily integrated to the existing learning-to-learn selection paradigms, to distinguish the hard samples from the mislabeled ones and further boost the performance of robust learning with noisy labels. Our NIB module consists of a probability transition matrix and an inter-class correlation (IC) loss. In particular, we calculate the Kullback-Leibler (KL) divergence between the network prediction and the accumulative soft-label generated from the probability transition matrix, which represents the inter-class correlation, as an auxiliary loss (\emph{i.e.,} IC loss). Such that, with lower value of IC loss, the hard cases can be easily distinguished from the ones with noisy labels. The proposed NIB module is evaluated on two publicly available natural and medical image datasets (CIFAR-10 and ISIC 2019). Experiemental results demonstrate that the proposed NIB significantly improves the classification accuracy of state-of-the-art robust training frameworks, \emph{e.g.,} Co-teaching and JoCoR.

\section{Method}

In this section, we first uncover the difference between hard and mislabeled samples from the aspect of inter-class correlation, and then present our noise ignoring block (NIB) in details.

\paragraph{Problem Formulation.}
We first illustrate the difference between hard samples and the ones with noisy labels.
Generally, the hard samples are the ones, which are mistakenly identified to a wrong but semantic-related class by the model, due to the similar visual features. For example, the dog contains more similar visual characteristics to cat than the other classes such as truck and airplane in CIFAR-10. Hence, the `hard' dog sample mistakenly classified to cat, \emph{i.e.,} falling around the decision boundary of dog/cat in the latent space, can aid the model to refine its decision boundary and should be paid more attention, which have been verified by existing studies \cite{Shrivastava_2016_CVPR, 9172136}. In contrast, the noisy labels are not necessary to be semantic-related, \emph{e.g.,} a dog sample with a noisy label of ship. Here, we reveal a common drawback shared by the conventional robust learning approaches---they calculate the cross-entropy loss between network predictions ($\bm{p}_{hard}$ and $\bm{p}_{noisy}$) and one-hot label $\bm{y}$, result in large loss values for both hard samples and the ones with noisy labels. Therefore, the valuable hard samples are often wrongly identified as mislabeled ones and excluded for network training by current robust learning frameworks, which degrades the performance of trained models. We argue that this issue is caused by the insufficient information provided by the one-hot label and propose that the soft label $\hat{\bm{y}}$, describing the inter-class relationship, is a potential solution for the problem.

\subsection{Noise Ignoring Block}

The proposed noise ignoring block (NIB) aims to mitigate negative effects of noisy labels by minimizing the distribution distance between soft labels and instance predictions. The pipeline of our NIB module can be divided into two processes, \emph{i.e.,} probability transition matrix estimation and inter-class correlation (IC) loss calculation, which are alternatively progressed, as presented in Alg.~\ref{alg:NIB}. The detailed information of each process is provided in the following:

\begin{algorithm}[!t]
\caption{Training Procedure with Noise Ignoring Block (NIB)}
\label{alg:NIB}
\begin{algorithmic}[1]
\REQUIRE Network $f$ with $\boldsymbol{\Theta}$, learning rate $\eta$, epoch number $N_e$, iteration number $N_i$, batch size $n_b$, training set $\mathcal{D} = \{ (\bm{x}_i, y_i) \}_{i=1}^{N}$ consists of $N$ samples from $K$ classes; 
\FOR{$n$ = 1,2,\ldots,$N_e$}
    \STATE \textbf{Shuffle} training set $\mathcal{D}$; 
    \FOR{$j= 1,\ldots,N_i$}
        \STATE \textbf{Fetch} mini-batch $\mathcal{D}_j$ with size $n_b$ from $\mathcal{D}$;
        \STATE $\bm{p}_{\bm{x}}$ = $f(\boldsymbol{x},\boldsymbol{\Theta})$, $\forall \boldsymbol{x} \in \mathcal{D}_j$;
        \STATE \textbf{Obtain} the potential set of `clean' samples $\mathcal{\Tilde{D}}_j$ with size $d$ from $\mathcal{D}_j$ based on the loss defined in Eq. (\ref{total loss}) using $\bm{p}_{\bm{x}}$, $\bm{y}$ and $\hat{\bm{y}}$ (obtained from $\mathbb{T}^{(n-1)}$);
        \FOR{$k= 1,\ldots,K$}
            \STATE \textbf{Update} $\bm{T}_k^{(n,j)}$ by (\ref{alg.8});
        \ENDFOR
        \STATE \textbf{Obtain} $\ell_{overall}$ by Eq. (\ref{total loss}) on $\mathcal{\Tilde{D}}_j$;
        \STATE \textbf{Update} $\boldsymbol{\Theta} = \boldsymbol{\Theta} - \eta \nabla \ell_{overall}$;
    \ENDFOR
        \STATE \textbf{Update} $\mathbb{T}^{(n)}$ by Eqs. (\ref{alg.10}) and (\ref{alg.15})
\ENDFOR
\ENSURE $\boldsymbol{\Theta}$
\end{algorithmic}
\end{algorithm}

\paragraph{Probability Transition Matrix.}
The probability transition matrix is formed by the accumulative soft-labels calculated using `clean' data, which represents the inter-class correlation. Denote $\mathcal{D} = \{ (\bm{x}_i, y_i) \}_{i=1}^{N}$ as the training set containing $N$ samples from $K$ classes, where $\bm{x}_i \in \mathcal{R}^{H\times W\times C}$ is the image ($H$, $W$ and $C$ are the height, width and number of channels of the image, respectively), and $y_i \in \{1,\ldots, K\}$ is the one-hot label. Let $\mathbb{T}$ indicate a $K\times K$ probability transition matrix,\footnote{The initial probability transition matrix is denoted as $\mathbb{T}^0 = \bm{0}_{K\times K}$ (\emph{i.e.,} a zero matrix).} which can be written as $\mathbb{T} = [\bm{T}_{1},\bm{T}_{2},\ldots,\bm{T}_{K}]$, where $\bm{T}_k$ ($k \in \{1,\ldots, K\}$) is the $k$-th row of $\mathbb{T}$.
After selecting `clean' samples $\mathcal{\Tilde{D}}_j$ from a batch $\mathcal{D}_j$,\footnote{At the beginning of robust learning, we select `clean' samples based on the cross-entropy loss calculated with the one-hot labels to generate $\mathbb{T}^1$. Then, the joint criterion defined in Eq.~\ref{total loss} is adopted for sample selection and matrix update. The number of clean data is consistent to the setting of \cite{han2018co, wei2020combating}} the calculation of $\bm{T}_k^{(n,j)}$ for class $k$ at current epoch $n$ can be formulated as:
\begin{equation}
    \label{alg.8}
    \bm{T}_k^{(n,j)}=\frac{1}{d_{k}} \sum_{(\bm{x}, y) \in \mathcal{\Tilde{D}}_{j,k}} \bm{p}_{\bm{x}},
\end{equation}
where $\mathcal{\Tilde{D}}_{j,k}=\{(\bm{x}, y) \mid y=k,\,\,\, \boldsymbol{x} \in \mathcal{\Tilde{D}}_j\}$; $\bm{p}_{\bm{x}}$ is the prediction (class-wise probability) yielded by the classification network $f$ with parameter $\boldsymbol{\Theta}$ for $\bm{x}$ ($\bm{p}_{\bm{x}}$ = $f(\boldsymbol{x},\boldsymbol{\Theta})$); and $d_k$ is the number of samples from class $k$ in the selected `clean' set $\mathcal{\Tilde{D}}_j$. After that, the probability transition matrix $\mathbb{T}^{(n,j)}$ for epoch $n$ and batch $\mathcal{D}_j$ is obtained via concatenation:
\begin{equation}
    \label{alg.10}
    \mathbb{T}^{(n,j)}= [\bm{T}_1^{(n,j)},\bm{T}_2^{(n,j)},\ldots,\bm{T}_K^{(n,j)}].
\end{equation}

The final probability transition matrix of epoch $n$ is obtained by averaging, as defined:
\begin{equation}
    \label{alg.15}
    \mathbb{T}^{(n)}=\frac{1}{n \times N_i} \sum_{m=1}^{n} \sum_{j=1}^{N_i} \mathbb{T}^{(m, j)},
\end{equation}
where $N_i = \frac {N} {n_b}$ is the number of iterations in one epoch ($n_b$ is the batch size).
The probability transition matrix $\mathbb{T}^{(n)}$ is used to calculate the inter-class correlation loss in the next epoch $n+1$ for the `clean' data selection. Concretely, $\bm{T}_k$ from $\mathbb{T}$ is used as the accumulative soft label $\hat{\bm{y}}$ for class $k$, \emph{i.e.,} $\hat{\bm{y}_i} = \bm{T}_k$ for $y_i=k$.

\paragraph{Inter-class Correlation Loss.}
To distinguish the `hard' samples from the mislabeled ones, we implement an inter-class correlation (IC) loss calculated with the soft labels from the estimated probability transition matrix $\mathbb{T}$. As previously mentioned, the accumulative soft label $\hat{\bm{y}}$ for class $k$ is derived from $\bm{T}_k$ of $\mathbb{T}$. Then, for an input image $\bm{x}_i$, the IC loss $\ell_{IC}$ can be formulated as:
\begin{equation}
    \label{soft loss}
    \ell_{IC}(\bm{x}_i)=D_{\mathrm{KL}}\left(\boldsymbol{\hat{y}}_i\| \bm{p}_{\bm{x}_i} \right)= \sum_{k=1}^{K} \hat{y}^{k}_i \log \frac{\hat{y}^{k}_i}{p^{k}\left(\boldsymbol{x}_{i}\right)},
\end{equation}
where $p^k(\bm{x}_i)$ is the $k$-th element of $\bm{p}_{\bm{x}_i}$ ; $\hat{y}^{k}_i$ is the $k$-th element of $\hat{\bm{y}}$; and $D_\mathrm{KL}$ is the KL divergence measuring the divergence between distributions.

\paragraph{Objective Function.} The proposed NIB is a plugin module, which can be easily integrated to the existing robust training framework. Hence, the overall criterion $\ell_{overall}$ for `clean' sample selection can be written as:
\begin{equation}
    \label{total loss}
    \ell_{overall}\left(\bm{x}_{i}\right)=\lambda \cdot \ell_{cls}\left(\bm{x}_{i}, y_{i}\right)+(1-\lambda) \cdot \ell_{IC}\left(\boldsymbol{x}_{i}, \hat{\bm{y}}_{i}\right),
\end{equation}
where $\ell_{cls}$ is the classification loss adopted by existing robust training approaches, \emph{e.g.,} cross-entropy loss in Co-teaching, and $\lambda$ is a factor balancing the $\ell_{cls}$ and $\ell_{IC}$ ($\lambda$ is empirically set to 0.6 in our experiments). Consistent to \cite{han2018co,wei2020combating}, the samples with lower $\ell_{overall}$ are selected as `clean' data by instance ranking \cite{han2018co,wei2020combating}. The $\ell_{overall}$ is not only used for `clean' data selection, but also for the optimization of classification network, as presented in Alg.~\ref{alg:NIB}.

\paragraph{\bf Relationship between Label Smoothing and Soft Label.}
Label smoothing is a technique widely used for the training of deep learning models, wherein one-hot training labels are mixed with uniform label vectors \cite{szegedy2016rethinking}. Empirically, label smoothing has been proven to improve both predictive performance and model calibration \cite{szegedy2016rethinking,zoph2018learning}. However, the label noises seem to be amplified by the label smoothing, since it is equivalent to injecting symmetric noise to the labels \cite{xie2016disturblabel}. In contrast, the soft label used in this study is a data-driven label, which is derived from the probability transition matrix estimated from the `clean' samples. In other words, the soft label is a representative of probability distribution corresponding to the class (\emph{vs.} the uniform distribution adopted by label smoothing); hence, no extra label noise is introduced to the dataset. This is also the underlying reason that the proposed inter-class correlation (IC) loss, calculated using the soft label and model prediction, can further benefit the robust learning with noisy labels. 

It is worthwhile to mention that soft label has been widely used for different tasks, \emph{e.g.,} semi-supervised learning \cite{Laine_2017_ICLR} and knowledge distillation \cite{Hinton_2014_NIPS}. Concretely, the former one \cite{Laine_2017_ICLR} accumulated the network predictions from different training epochs as the supervision signal (soft label) for unlabeled data. The later one \cite{Hinton_2014_NIPS} used the class-wise soft label for knowledge distillation. To this end, the proposed NIB module can be seen as the combination of the two approaches, which accumulates the network predictions from different training epochs to form a class-wise transition matrix for noisy label rejection. Furthermore, we notice that there are some existing approaches \cite{Meta_2020} using the transition matrix for robust network training. The related noisy label defencing framework \cite{Meta_2020} was a typical meta-learning-based approach, which required a meta net to iteratively optimize the parameter of classifier. In contrast, the proposed NIB is a plug-in module without extra network parameters, which is flexible and easy-to-implement.

\section{Experiments}
In this section, we validate the proposed plugin module (noise ignoring block, NIB) on two publicly available natural and medical image datasets, and present the experimental results.

\subsection{Datasets} 
\paragraph{\bf CIFAR-10.} CIFAR-10\footnote{\url{https://www.cs.toronto.edu/~kriz/cifar.html}} \cite{cifar} is popularly used for evaluation of noisy labels in the literature \cite{goldberger2016training,patrini2017making,Quan2020miccai,reed2014training}. The dataset contains 60,000 images, which can be categorized to ten classes, with a uniform size of 32 $\times$ 32 pixels. The training set and test set consist of 50,000 images and 10,000 images, respectively.

\paragraph{\bf ISIC 2019.} Deep-learning-based medical image classification frameworks often suffer from the noisy label problem, due to the different experience levels of annotators (\emph{i.e.,} doctors and radiologists). In this paper, the widely-used ISIC 2019 dataset\footnote{\url{https://challenge2019.isic-archive.com/}} is adopted to verify the effectiveness of our NIB module for medical image classification.
The ISIC 2019 dataset \cite{tschandl2018ham10000} is from the challenge of prediction of eight skin disease categories with dermoscopic images, including melanoma (MEL), melanocytic nevus (NV), basal cell carcinoma (BCC), actinic keratosis (AK), benign keratosis (BKL), dermatofibroma (DF), vascular lesion (VASC), and squamous cell carcinoma (SCC). The original ISIC dataset is highly imbalanced between classes \cite{Continual_Dual_Distillation}. To alleviate the potential effect of imbalance in continual experiment, we randomly sample 628 images from each class (Note that we take all images from two classes with fewer than 628 images), consistent to \cite{Continual_Dual_Distillation}.
A total of 4,260 images are randomly divided into a training and a test set according to the ratio of 80:20.

\subsection{Experimental Settings} 
\paragraph{\bf Label Shuffling.} Following \cite{patrini2017making,reed2014training}, we shuffle the labels of the training set by a noise transition matrix $Q$, where $Q_{ij} = Pr[\Tilde{y}=j| y=i]$ denotes the probability of flipping class $i$ to $j$.
The widely-used structures of $Q$, \emph{i.e.,} symmetry flipping \cite{van2015learning,Quan2020miccai} and pair flipping \cite{han2018co}, are adopted in our study. Note that, consistent to \cite{van2015learning,Quan2020miccai,han2018co}, we validate our NIB module with different noise ratios, denoting as `symmetry-10\%', `symmetry-20\%', `symmetry-40\%' and `pair-10\%'. For example, the `symmetry-10\%' represents that 10\% of the labels have been symmetrically flipped to be noisy labels.

\paragraph{\bf Implementation \& Evaluation Criterion.} The proposed NIB is implemented using the PyTorch toolbox. All the frameworks use the same backbone architectures, \emph{i.e.,} a 9-layer CNN network architecture is adopted for CIFAR-10, while we utilize the ResNet-18 and DenseNet-169 as the backbone for ISIC 2019, due to the larger image size and classification complexity of the ISIC dataset. The Adam optimizer (momentum=0.9) is used for network optimization with an initial learning rate of 0.001. The batch size is set to 128 and 64 for CIFAR-10 and ISIC dataset, respectively. The images from ISIC 2019 dataset are resized to $224\times224$ pixels for network processing. The average classification accuracy (ACC) on the test set is adopted as the metric to evaluate the performance of robust learning with noisy labels. We run 200 epochs in total and calculate ACC over the last 10 epochs.

\subsection{Performance Evaluation}
In this section, we present the experimental results on the two publicly available datasets, \emph{i.e,} CIFAR-10 and ISIC 2019. The state-of-the-art robust training approaches, \emph{i.e.,} Co-teaching \cite{han2018co} and JoCoR \cite{wei2020combating}, are involved as baselines.

\paragraph{\bf CIFAR-10.} The average classification accuracy on the CIFAR-10 test set yielded by the classification networks trained with different strategies is listed in Table~\ref{cifar10}. It can be observed that the proposed NIB module consistently improves the classification accuracy of existing robust learning approaches under different ratios of noises. Concretely, even when 40\% of the labels are symmetrically flipped (\emph{i.e.,} symmetry-40\%), our NIB can still significantly increase the test ACC for Co-teaching and JoCoR by margins of $+4.73\%$ and $+4.55\%$, respectively, which demonstrates the outstanding robustness of our NIB to label noises.

As shown in Table~\ref{cifar10}, we also conduct an ablation study on CIFAR-10 to evaluate the performance of models only using $\ell_{IC}$ for clean data selection. As the sample-wise class relationship information may differ from the class-wise one, which results in a larger value of $\ell_{IC}$, some clean data is wrongly rejected while only using $\ell_{IC}$ for clean data selection. Hence, the classification accuracy of models only using $\ell_{IC}$ significantly degrades on CIFAR-10, compared to the original learning-to-learn paradigms using $\ell_{cls}$. The experimental results also reveal the mechanism underlying our framework---the clean data is easily filtered by $\ell_{cls}$, while $\ell_{IC}$ is proposed to further separate the hard samples from the ones with noisy labels.
In addition, we integrate the state-of-the-art approach (dynamic bootstrapping, $\ell_{DY}$)\cite{pmlr-v97-arazo19a} to existing learning-to-learn paradigms for comparison. As shown in Table~\ref{cifar10}, the frameworks using our NIB module surpass the ones using $\ell_{DY}$ by a large margin.

\paragraph{\bf ISIC 2019.} To validate the effectiveness of our NIB module on medical images, we conduct experiments on the ISIC 2019 dataset. The test ACC on the ISIC 2019 dataset are presented in Table \ref{ISIC}. A similar trend to CIFAR-10 is observed---using our NIB module, the ACCs of existing approaches on the ISIC 2019 test set are consistently improved. The Co-teaching + NIB achieves the best test ACC under most noise ratios. Furthermore, we notice that the proposed NIB achieves significant improvements for not only ResNet-18, but also the ultra-deep DenseNet-169, even under the large noise ratio (40\%).

\begin{table}[!t]
    \centering
    \caption{Average classification accuracy (\%) on the CIFAR-10 test set.}\label{cifar10}
    \small
    \scalebox{0.9}{
    \begin{tabular}{c|c |c c c| c}
    \hline
    Method$\downarrow$,Noise Type$\rightarrow$  & Clean          & symmetry-10\%  & symmetry-20\%  & symmetry-40\%  & pair-10\%         \\\hline
    Co-teaching \cite{han2018co}                & 89.13          & 85.06          & 82.24          & 78.15          & 85.29             \\
    Co-teaching+NIB                             & \textbf{90.87} & \textbf{88.44} & \textbf{86.69} & \textbf{82.88} & \textbf{88.33}    \\
    Co-teaching+$\ell_{IC}$-only                & 90.13              & 87.80          & 86.07          & 76.92          & 86.34             \\
    Co-teaching+$\ell_{DY}$\cite{pmlr-v97-arazo19a}     & 89.04              & 85.23	  &83.32  	&79.93	
    &83.80    \\\hline
    JoCoR \cite{wei2020combating}               & 89.30          & 84.04          & 81.61          & 77.40          & 84.40             \\
    JoCoR+NIB                                   & \textbf{90.70} & \textbf{87.81} & \textbf{85.88} & \textbf{81.95} & \textbf{87.61}    \\
    JoCoR+$\ell_{IC}$-only                      & 90.69              & 79.39          & 68.88          & 64.84          & 70.73             \\
    JoCoR+$\ell_{DY}$\cite{pmlr-v97-arazo19a}           & 90.16              & 82.36	  &82.44	    &79.67	
    &84.20\\\hline
    \end{tabular}}
\end{table}

\begin{table}[!t]
\caption{Average test ACC (\%) on the ISIC 2019 dataset.}\label{ISIC}
\small
\centering
\scalebox{0.8}{
    \begin{tabular}{c|c|c|c c c|c}
    \hline
    Backbone                      & Method$\downarrow$,Noise Type$\rightarrow$ & Clean          & Symmetry-10\%  & Symmetry-20\%   & Symmetry-40\%  & Pair-10\%      \\ \hline
    \multirow{4}{*}{ResNet-18}    & Co-teaching \cite{han2018co}      & 62.00          & 58.47          & 55.43           & 47.81          & 58.40          \\
                                  & Co-teaching+NIB   & \textbf{64.40} & \textbf{60.08} & \textbf{59.37}  & \textbf{49.33} & \textbf{60.08} \\ \cline{2-7} 
                                  & JoCoR \cite{wei2020combating}            & 62.11         & 57.78          & 56.71           & 47.62          & 58.67          \\
                                  & JoCoR+NIB         & \textbf{63.43} & \textbf{59.77} & \textbf{59.17}  & \textbf{49.65} & \textbf{60.43} \\ \hline
    \multirow{4}{*}{DenseNet-169} & Co-teaching  & 70.11         & 65.67          & 63.29           & 48.87          & 65.16        \\
                                  & Co-teaching+NIB   & \textbf{71.32} & \textbf{68.86} & \textbf{63.74} & \textbf{56.12} & \textbf{68.11} \\ \cline{2-7} 
                                  & JoCoR           & 68.46          & 62.93          & 56.08           & 49.92          & 67.24          \\
                                  & JoCoR+NIB         & \textbf{70.99} & \textbf{65.79} & \textbf{60.07}  & \textbf{53.42} & \textbf{68.22} \\ \hline
    \end{tabular}}
\end{table}

\begin{table}[!t]
    \centering
    \small
    \caption{Classification accuracy (\%) on the Clothing1M test set.}\label{clothing}
    \begin{tabular}{l|c|c|c|c}
    \hline
    Method      & Co-teaching \cite{han2018co} & Co-teaching+NIB & JoCoR \cite{wei2020combating} & JoCoR+NIB      \\\hline
    Accuracy    & 69.74                        & \textbf{71.02}  &70.25                          & \textbf{71.28} \\\hline
    \end{tabular}
\end{table}

\paragraph{\bf Application on Real-world Dataset.} An experiment is conducted on the Clothing1M\footnote{\url{https://github.com/Cysu/noisy_label}} \cite{Xiao_2015_CVPR}, a real-world noisy dataset, to further validate the effectiveness of our NIB module. The dataset consists of one million images captured from online shopping websites. The label (14 classes) for each image is generated by extracting tags from the surronding texts and keywords, which are naturally noisy. The Clothing1M dataset is separated to training and test sets according to the protocol \cite{wei2020combating}. In the experiment, frameworks are trained with noisy training set, and then evaluated on the test set using clean labels. The evaluation results are shown in Table~\ref{clothing}. It can be observed that the proposed NIB module consistently boosts the classification accuracy of existing learning-to-learn paradigms, \emph{i.e.,} improvements of $+1.28\%$ and $+1.03\%$ are yielded by our NIB module to Co-teaching and JoCoR, respectively. The experimental results demonstrate the effectiveness of the proposed NIB module for the realistic application.

\paragraph{Label Precision.} Apart from ACC, the label precision is also measured and reported on CIFAR-10 and ISIC 2019 datasets to assess the pure ratio of selected samples: 

\begin{equation}
    \textsl{label precision} = \frac {\textsl{the number of selected samples with correct labels}} {\textsl{the total number of selected samples}}.
\end{equation}
Intuitively, the higher label precision means the fewer noisy instances mistakenly identified as `clean' by the approach, \emph{i.e.,} the better robustness to noisy labels. 

\begin{figure}[!t]
    \begin{center}
    \includegraphics[width=0.9\linewidth]{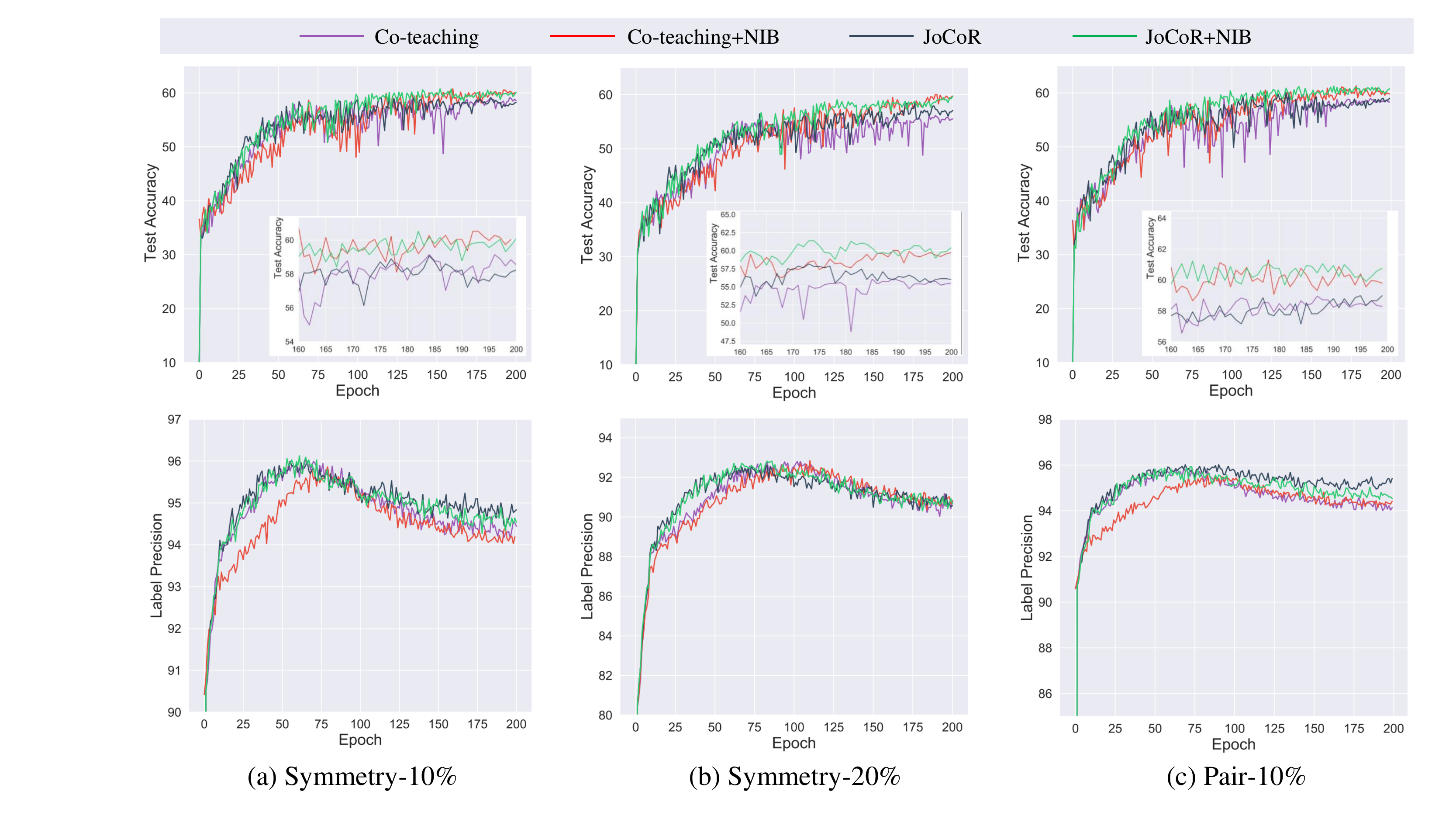}
    \caption{Training curves of {\bf ResNet-18} on the ISIC dataset, respectively. Top: test ACC (\%) \emph{vs.} epochs; bottom: label precision (\%) \emph{vs.} epochs. The figures on the top row reveal that the test accuracy of Co-teaching/JoCoR + NIB is consistently higher than the original ones in the late stage of training (\emph{i.e., } after 160 epochs of training).} \label{line_skin_res_resnet18}
    \end{center}
\end{figure}

\begin{figure}[!t]
    \begin{center}
    \includegraphics[width=0.9\linewidth]{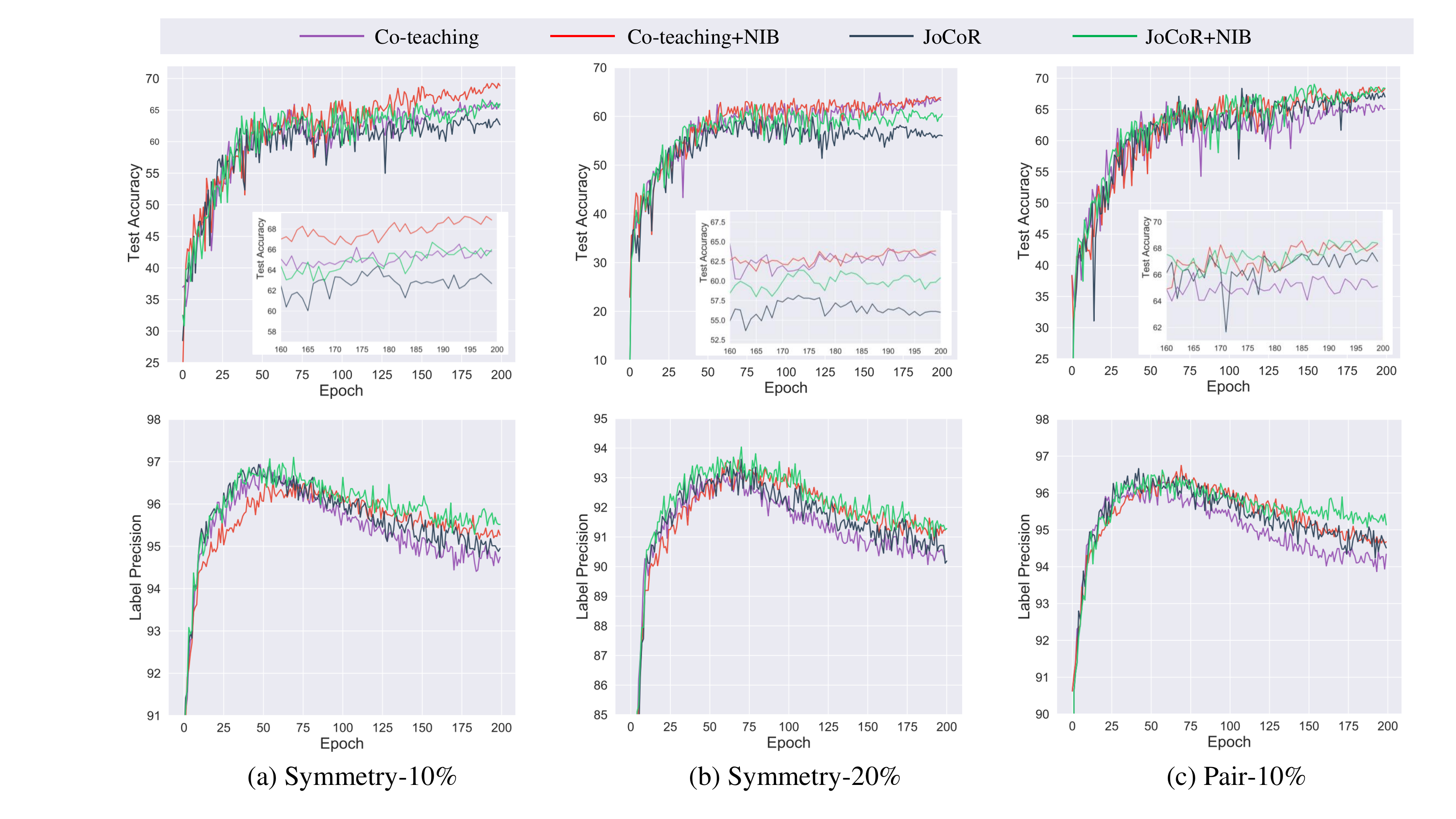}
    \caption{Training curves of {\bf DenseNet-169} (right) on the ISIC dataset, respectively. Top: test ACC (\%) \emph{vs.} epochs; bottom: label precision (\%) \emph{vs.} epochs. The figures on the top row reveal that the test accuracy of Co-teaching/JoCoR + NIB is consistently higher than the original ones in the late stage of training (\emph{i.e., } after 160 epochs of training).} \label{line_skin_res_densenet169}
    \end{center}
\end{figure}

The label precision of different robust learning approaches on CIFAR-10 and ISIC 2019 are listed in Table~\ref{cifar10-pre} and \ref{ISIC-pre}, respectively. The label precision of the frameworks using the proposed NIB module is observed to consistently surpass the original ones on the CIFAR-10 dataset. 
For the ISIC 2019 dataset, as stated in recent studies \cite{zhang2019understanding}, the ultra-deep networks (\emph{e.g.,} DenseNet-169) more easily overfit to the noises and severely suffer from the noisy label problem, compared to the shallow ones. 
Table~\ref{ISIC} reveals that the proposed NIB module can significantly alleviate such an overfitting problem, \emph{i.e.,} further improving the test accuracy of DenseNet-169 even under the large noise ratio (40\%). The underlying reason is the higher label precision achieved by our NIB module, as shown in Table \ref{ISIC-pre}.
The samples selected by the frameworks with our NIB is `cleaner' than the conventional ones, \emph{i.e.,} a higher label precision is achieved by Co-teaching + NIB and JoCoR + NIB than that of Co-teaching and JoCoR, which accordingly benefits the training of ultra-deep networks on the classification task. On the other hand, we notice that although the label precision of JoCoR + NIB is slightly lower than JoCoR in Table \ref{ISIC-pre} for ResNet-18, the test accuracy is still improved by using our NIB. This is because the JoCoR + NIB is joint optimized by $\ell_{cls}$ and $\ell_{IC}$. The combination of these two losses refines the direction of optimization, compared to the $\ell_{cls}$-only JoCoR. The curves of test accuracy and label precision \emph{vs.} epochs during the whole training process of ResNet-18 and DenseNet-169 on the ISIC dataset are presented in Fig.~\ref{line_skin_res_resnet18} and Fig.~\ref{line_skin_res_densenet169}. It can be observed that the frameworks using our NIB consistently outperform the original ones on both test accuracy and label precision, demonstrating the generalization of the proposed NIB module.

\begin{table}[!t]
    \caption{Average label precision (\%) on the CIFAR-10 over the last 10 epochs.}\label{cifar10-pre}
    \small
    \centering
    \begin{tabular}{c|c c c|c}
    \hline
    Method$\downarrow$,Noise Type$\rightarrow$  & Symmetry-10\%             & Symmetry-20\%        & Symmetry-40\% & Pair-10\% \\ \hline
    Co-teaching \cite{han2018co}       & 95.32                     & 92.74                &      87.46         & 94.94     \\
    Co-teaching+NIB   & \textbf{96.01}               &    \textbf{93.43}                   &    \textbf{88.28}           & \textbf{95.73}     \\ \hline
    JoCoR \cite{wei2020combating}            & 95.76                     & 93.43                & 89.06         & 95.29     \\
    JoCoR+NIB         & \textbf{96.23}            & \textbf{93.90}              & \textbf{89.57}     & \textbf{96.01}     \\ \hline
    \end{tabular}
\end{table}

\begin{table}[!t]
    \caption{Label precision (\%) of the selected samples on the ISIC 2019 dataset.}\label{ISIC-pre}
    \centering
    \scalebox{0.8}{
    \begin{tabular}{c|c|c c c|c}
    \hline
    Backbone                 & Method$\downarrow$,Noise Type$\rightarrow$       & Symmetry-10\% & Symmetry-20\% & Symmetry-40\% & Pair-10\% \\ \hline
    \multirow{4}{*}{ResNet-18}                              & Co-teaching             & \textbf{94.33}       & 90.58       & 80.76       & 94.16  \\
                                 & Co-teaching+NIB & 94.17       & \textbf{90.79}       & \textbf{82.03}       & \textbf{94.31}   \\ \cline{2-6} 
                               & JoCoR                   & \textbf{94.69}         & \textbf{90.77}       & \textbf{81.64}         & \textbf{95.26}     \\
       & JoCoR+NIB      & 94.55         & 90.74       & 81.04              & 94.65     \\ \hline
    \multirow{4}{*}{DenseNet-169}  & Co-teaching             & 94.76         & 90.41         & 78.42         & 94.17     \\
                                 & Co-teaching+NIB & \textbf{95.30}          & \textbf{91.18}         & \textbf{82.06}         & \textbf{94.71}     \\ \cline{2-6} 
                                & JoCoR                   & 94.98         & 90.65         & 79.84         & 94.63     \\
     & JoCoR+NIB      & \textbf{95.62}         & \textbf{91.35}         &  \textbf{81.73}            & \textbf{95.27}     \\ \hline
    \end{tabular}}
\end{table}


\begin{figure}[!tb]
\centering
	\begin{minipage}[t]{0.16\textwidth}
		\centering
		\includegraphics[width=0.7\textwidth]{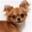}
		\footnotesize{\\(a) dog\\(hard case 1)\\ $\ell_{cls}: 1.47$\\ $\ell_{IC}: \bm{1.63}$}
	\end{minipage}
	\begin{minipage}[t]{0.16\textwidth}
		\centering
		\includegraphics[width=0.7\textwidth]{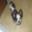}
		\footnotesize{\\(b) dog\\(hard case 2)\\ $\ell_{cls}: 1.38$\\  $\ell_{IC}: \bm{1.45}$}
	\end{minipage}
	\begin{minipage}[t]{0.16\textwidth}
		\centering
		\includegraphics[width=0.7\textwidth]{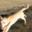}
		\footnotesize{\\(c) dog\\(hard case 3)\\ $\ell_{cls}: 1.56$\\  $\ell_{IC}: \bm{1.46}$}
	\end{minipage}
	\begin{minipage}[t]{0.16\textwidth}
		\centering
		\includegraphics[width=0.7\textwidth]{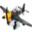}
		\footnotesize{\\(d) airplane\\(noisy label: ship)\\ $\ell_{cls}: 1.43$\\ $\ell_{IC}: \bm{2.66}$}
	\end{minipage}
	\begin{minipage}[t]{0.16\textwidth}
		\centering
		\includegraphics[width=0.7\textwidth]{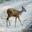}
		\footnotesize{\\(e) deer\\(noisy label: dog)\\ $\ell_{cls}: 1.52$\\  $\ell_{IC}: \bm{2.71}$}
	\end{minipage}
	\begin{minipage}[t]{0.16\textwidth}
		\centering
		\includegraphics[width=0.7\textwidth]{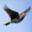}
		\footnotesize{\\(f) bird\\(noisy label: ship)\\ $\ell_{cls}: 1.38$\\  $\ell_{IC}: \bm{2.89}$}
	\end{minipage}
	\caption{Examples of hard and mislabeled cases from CIFAR-10 identified by our NIB.} \label{hard_noisy_samples_CIFAR10}
\end{figure}

\begin{figure}[!tb]
\centering
	\begin{minipage}[t]{0.2\textwidth}
		\centering
		\includegraphics[width=\textwidth]{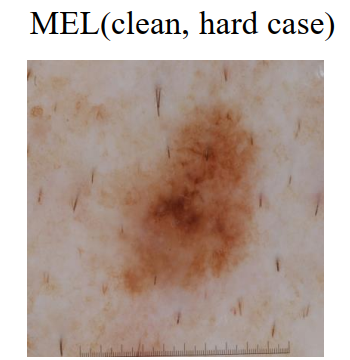}
		\footnotesize{\\(a) Hard case 1\\ $\ell_{cls}: 1.61$\\ $\ell_{IC}: \bm{1.60}$}
	\end{minipage}
	\begin{minipage}[t]{0.2\textwidth}
		\centering
		\includegraphics[width=\textwidth]{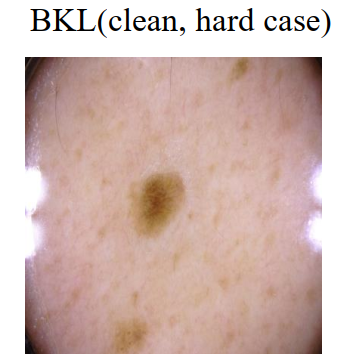}
		\footnotesize{\\(b) Hard case 2\\ $\ell_{cls}: 1.89$\\  $\ell_{IC}: \bm{1.45}$}
	\end{minipage}
	\begin{minipage}[t]{0.2\textwidth}
		\centering
		\includegraphics[width=\textwidth]{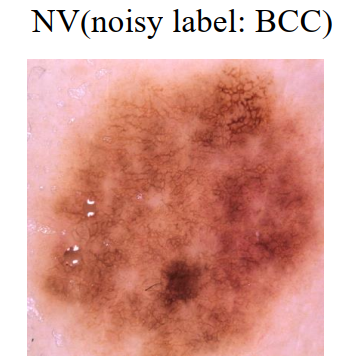}
		\footnotesize{\\(c) Noisy Label case 1\\ $\ell_{cls}: 1.61$\\ $\ell_{IC}: \bm{2.82}$}
	\end{minipage}
	\begin{minipage}[t]{0.2\textwidth}
		\centering
		\includegraphics[width=\textwidth]{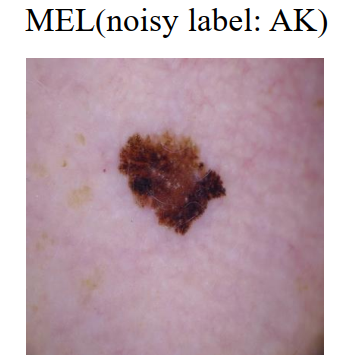}
		\footnotesize{\\(d) Noisy Lable case 2\\ $\ell_{cls}: 1.77$\\  $\ell_{IC}: \bm{2.74}$}
	\end{minipage}
	\caption{Examples of hard and mislabeled cases from ISIC 2019 identified by our NIB.} \label{hard_noisy_samples}
\end{figure}

\subsection{Analysis: Hard Samples \emph{vs.} Samples with Noisy Labels} \label{sec:sample_comparison}
In this section, to validate the effectiveness of inter-class correlation loss for hard sample identification, we show the detected hard samples and samples with noisy labels from CIFAR-10 and ISIC 2019 in Fig.~\ref{hard_noisy_samples_CIFAR10} and \ref{hard_noisy_samples}, respectively. The $\ell_{cls}$ and $\ell_{IC}$ calculated with one-hot label $\bm{y}$ and accumulative soft label $\hat{\bm{y}}$, respectively, are also listed under each sample. We observe that the $\ell_{cls}$ of hard and mislabeled cases is close, \emph{i.e.,} around $1.5$ and $1.7$ on CIFAR-10 and ISIC 2019, respectively. Therefore, the existing $\ell_{cls}$-only robust learning frameworks cannot distinguish them and will assign them to the same category (either `clean' or `noisy'). Thus, more samples with noise labels are involved for classifier training if treating them as `clean' or the hard samples are over filtered, both leading to the degradation of classification performance.
In contrast, our inter-class correlation loss $\ell_{IC}$ excellently distinguishes the hard samples and the ones with noisy labels (around 1.5 \emph{vs.} around 2.8) on both CIFAR-10 and ISIC datasets. Hence, the framework using our $\ell_{IC}$ can simultaneously benefit the classifier from hard samples and maintain the robustness to noise labels, which results in the improvements presented in the previous section.

\paragraph{\bf Effectiveness of Hard Samples for Network Training.} As aforementioned, the hard samples falling around the decision boundary in the latent space may provide rich information for decision boundary refinement and should be paid more attention during network training. Such a claim has been verified by the existing studies \cite{9172136, Shrivastava_2016_CVPR}. In our experiments, the experimental results explicitly demonstrate the effectiveness of hard samples for network training. The original learning-to-learn paradigms (Co-teaching and JoCoR) exclude the hard samples illustrated in Fig.~\ref{hard_noisy_samples_CIFAR10} and \ref{hard_noisy_samples} from network training, due to the large values of $\ell_{cls}$. Note that $\ell_{cls}$ is used as the overall loss $\ell_{overall}$ in the original paradigms for sample selection and the average $\ell_{cls}$ and $\ell_{IC}$ for clean samples are $1.13$ and $1.37$, respectively. In contrast, our $\ell_{IC}$ decreases the $\ell_{overall}$ (\emph{i.e.,} $\ell_{cls} + \ell_{IC}$) of hard samples; therefore, those samples can be separated from noisy ones and included for network training. As shown in Table~\ref{cifar10} and \ref{ISIC}, the classification accuracy is significantly improved by adding our $\ell_{IC}$ loss to $\ell_{overall}$, which validates the effectiveness of including hard samples for network training.

\section{Conclusion}
In this paper, we proposed a simple-yet-effective plugin module, namely noise ignoring block (NIB), which can be easily integrated to the existing learning-to-learn instance selection paradigms, to distinguish the hard samples from the mislabeled ones and further boost the performance of robust learning with noisy labels. Extensive experiments were conducted on natural and medical image datasets (CIFAR-10 and ISIC 2019). The experimental results showed that our NIB module consistently improved the performances of the state-of-the-art robust training methods, \emph{e.g.,} Co-teaching and JoCoR.


\paragraph{\bf Limitation and Future Work.} Currently, the proposed NIB module is only integrated to the framework designed for image classification. However, existing studies began to focus on the multi-label classification task, \emph{e.g.,} ReLabel \cite{Yun_2021_CVPR}. In this regard, we plan to adjust the proposed NIB module to more image processing tasks, such as multi-label classification and semantic segmentation, in the future work.

\section*{Acknowledgements}
This work was founded by the Key-Area Research and Development Program of Guangdong Province, China (No.2020B090923003 and No.2018B010111001), National Natural Science Foundation of China (Grant No. 52075285), National Key R\&D Program of China (2018YFC-2000702) and the Scientific and Technical Innovation 2030-``New Generation Artificial Intelligence'' Project (No. 2020AAA0104100).

\bibliography{main}

\begin{thebibliography}{29}
\providecommand{\natexlab}[1]{#1}
\providecommand{\url}[1]{\texttt{#1}}
\expandafter\ifx\csname urlstyle\endcsname\relax
  \providecommand{\doi}[1]{doi: #1}\else
  \providecommand{\doi}{doi: \begingroup \urlstyle{rm}\Url}\fi

\bibitem[Arazo et~al.(2019)Arazo, Ortego, Albert, O'Connor, and
  Mcguinness]{pmlr-v97-arazo19a}
Eric Arazo, Diego Ortego, Paul Albert, Noel O'Connor, and Kevin Mcguinness.
\newblock Unsupervised label noise modeling and loss correction.
\newblock In \emph{Proceedings of the International Conference on Machine
  Learning (ICML)}, 2019.

\bibitem[Goldberger and Ben-Reuven(2016)]{goldberger2016training}
Jacob Goldberger and Ehud Ben-Reuven.
\newblock Training deep neural-networks using a noise adaptation layer.
\newblock In \emph{Proceedings of the 5th International Conference on Learning
  Representation}, 2016.

\bibitem[Han et~al.(2018)Han, Yao, Yu, Niu, Xu, Hu, Tsang, and
  Sugiyama]{han2018co}
Bo~Han, Quanming Yao, Xingrui Yu, Gang Niu, Miao Xu, Weihua Hu, Ivor Tsang, and
  Masashi Sugiyama.
\newblock Co-teaching: Robust training of deep neural networks with extremely
  noisy labels.
\newblock \emph{arXiv preprint arXiv:1804.06872}, 2018.

\bibitem[Hinton et~al.(2014)Hinton, Vinyals, and Dean]{Hinton_2014_NIPS}
Geoffrey Hinton, Oriol Vinyals, and Jeff Dean.
\newblock Distilling the knowledge in a neural network.
\newblock \emph{arXiv preprint arXiv:1503.02531}, 2014.

\bibitem[Jiang et~al.(2018)Jiang, Zhou, Leung, Li, and
  Fei-Fei]{jiang2018mentornet}
Lu~Jiang, Zhengyuan Zhou, Thomas Leung, Li-Jia Li, and Li~Fei-Fei.
\newblock {MentorNet}: Learning data-driven curriculum for very deep neural
  networks on corrupted labels.
\newblock In \emph{International Conference on Machine Learning}, pages
  2304--2313, 2018.

\bibitem[Krizhevsky(2009)]{cifar}
Alex Krizhevsky.
\newblock \emph{Learning multiple layers of features from tiny images}, 2009.

\bibitem[Laine and Aila(2017)]{Laine_2017_ICLR}
Samuli Laine and Timo Aila.
\newblock Temporal ensembling for semi-supervised learning.
\newblock In \emph{International Conference on Learning Representations
  (ICLR)}, 2017.

\bibitem[Li et~al.(2020{\natexlab{a}})Li, Wei, Chen, Cao, Zhou, Zhu, Wu, Lan,
  Sun, Qian, Ma, Xu, and Zheng]{9172136}
Yuexiang Li, Dong Wei, Jiawei Chen, Shilei Cao, Hongyu Zhou, Yanchun Zhu,
  Jianrong Wu, Lan Lan, Wenbo Sun, Tianyi Qian, Kai Ma, Haibo Xu, and Yefeng
  Zheng.
\newblock Efficient and effective training of covid-19 classification networks
  with self-supervised dual-track learning to rank.
\newblock \emph{IEEE Journal of Biomedical and Health Informatics}, 24\penalty0
  (10):\penalty0 2787--2797, 2020{\natexlab{a}}.

\bibitem[Li et~al.(2020{\natexlab{b}})Li, Zhong, Wang, and
  Zheng]{Continual_Dual_Distillation}
Zhuoyun Li, Changhong Zhong, Ruixuan Wang, and Wei-Shi Zheng.
\newblock Continual learning of new diseases with dual distillation and
  ensemble strategy.
\newblock In \emph{Medical Image Computing and Computer Assisted Intervention},
  pages 169--178, 2020{\natexlab{b}}.
\newblock ISBN 978-3-030-59710-8.

\bibitem[Liu and Tao(2015)]{liu2015classification}
Tongliang Liu and Dacheng Tao.
\newblock Classification with noisy labels by importance reweighting.
\newblock \emph{IEEE Transactions on Pattern Analysis and Machine
  Intelligence}, 38\penalty0 (3):\penalty0 447--461, 2015.

\bibitem[Malach and Shalev-Shwartz(2017)]{malach2017decoupling}
Eran Malach and Shai Shalev-Shwartz.
\newblock Decoupling ``when to update" from ``how to update".
\newblock \emph{arXiv preprint arXiv:1706.02613}, 2017.

\bibitem[Masi et~al.(2018)Masi, Wu, Hassner, and Natarajan]{masi2018deep}
Iacopo Masi, Yue Wu, Tal Hassner, and Prem Natarajan.
\newblock Deep face recognition: A survey.
\newblock In \emph{2018 31st SIBGRAPI Conference on Graphics, Patterns and
  Images}, pages 471--478. IEEE, 2018.

\bibitem[Menon et~al.(2015)Menon, Van~Rooyen, Ong, and
  Williamson]{menon2015learning}
Aditya Menon, Brendan Van~Rooyen, Cheng~Soon Ong, and Bob Williamson.
\newblock Learning from corrupted binary labels via class-probability
  estimation.
\newblock In \emph{International Conference on Machine Learning}, pages
  125--134, 2015.

\bibitem[Patrini et~al.(2017)Patrini, Rozza, Krishna~Menon, Nock, and
  Qu]{patrini2017making}
Giorgio Patrini, Alessandro Rozza, Aditya Krishna~Menon, Richard Nock, and
  Lizhen Qu.
\newblock Making deep neural networks robust to label noise: A loss correction
  approach.
\newblock In \emph{Proceedings of the IEEE Conference on Computer Vision and
  Pattern Recognition}, pages 1944--1952, 2017.

\bibitem[Quan et~al.(2020)Quan, Li, Chen, and Zhang]{Quan2020miccai}
Li~Quan, Yan Li, Xiaoyi Chen, and Ni~Zhang.
\newblock An effective data refinement approach for upper gastrointestinal
  anatomy recognition.
\newblock In \emph{International Conference on Medical Image Computing and
  Computer Assisted Intervention}, pages 43--52, 2020.

\bibitem[Reed et~al.(2014)Reed, Lee, Anguelov, Szegedy, Erhan, and
  Rabinovich]{reed2014training}
Scott Reed, Honglak Lee, Dragomir Anguelov, Christian Szegedy, Dumitru Erhan,
  and Andrew Rabinovich.
\newblock Training deep neural networks on noisy labels with bootstrapping.
\newblock \emph{arXiv preprint arXiv:1412.6596}, 2014.

\bibitem[Sakai et~al.(2017)Sakai, Plessis, Niu, and Sugiyama]{sakai2017semi}
Tomoya Sakai, Marthinus~Christoffel Plessis, Gang Niu, and Masashi Sugiyama.
\newblock Semi-supervised classification based on classification from positive
  and unlabeled data.
\newblock In \emph{International Conference on Machine Learning}, pages
  2998--3006, 2017.

\bibitem[Shrivastava et~al.(2016)Shrivastava, Gupta, and
  Girshick]{Shrivastava_2016_CVPR}
Abhinav Shrivastava, Abhinav Gupta, and Ross Girshick.
\newblock Training region-based object detectors with online hard example
  mining.
\newblock In \emph{Proceedings of the IEEE Conference on Computer Vision and
  Pattern Recognition (CVPR)}, pages 761--769, 2016.

\bibitem[Shu et~al.(2020)Shu, Zhao, Xu, and Meng]{Meta_2020}
Jun Shu, Qian Zhao, Zongben Xu, and Deyu Meng.
\newblock Meta transition adaptation for robust deep learning with noisy
  labels.
\newblock \emph{arXiv preprint arXiv:2006.05697}, 2020.

\bibitem[Szegedy et~al.(2016)Szegedy, Vanhoucke, Ioffe, Shlens, and
  Wojna]{szegedy2016rethinking}
Christian Szegedy, Vincent Vanhoucke, Sergey Ioffe, Jon Shlens, and Zbigniew
  Wojna.
\newblock Rethinking the inception architecture for computer vision.
\newblock In \emph{Proceedings of the IEEE Conference on Computer Vision and
  Pattern Recognition}, pages 2818--2826, 2016.

\bibitem[Tan et~al.(2020)Tan, Pang, and Le]{tan2020efficientdet}
Mingxing Tan, Ruoming Pang, and Quoc~V Le.
\newblock {EfficientDet}: Scalable and efficient object detection.
\newblock In \emph{Proceedings of the IEEE/CVF Conference on Computer Vision
  and Pattern Recognition}, pages 10781--10790, 2020.

\bibitem[Tschandl et~al.(2018)Tschandl, Rosendahl, and
  Kittler]{tschandl2018ham10000}
Philipp Tschandl, Cliff Rosendahl, and Harald Kittler.
\newblock The {HAM10000} dataset, a large collection of multi-source
  dermatoscopic images of common pigmented skin lesions.
\newblock \emph{Scientific Data}, 5\penalty0 (1):\penalty0 1--9, 2018.

\bibitem[Van~Rooyen et~al.(2015)Van~Rooyen, Menon, and
  Williamson]{van2015learning}
Brendan Van~Rooyen, Aditya~Krishna Menon, and Robert~C Williamson.
\newblock Learning with symmetric label noise: The importance of being
  unhinged.
\newblock \emph{arXiv preprint arXiv:1505.07634}, 2015.

\bibitem[Wei et~al.(2020)Wei, Feng, Chen, and An]{wei2020combating}
Hongxin Wei, Lei Feng, Xiangyu Chen, and Bo~An.
\newblock Combating noisy labels by agreement: A joint training method with
  co-regularization.
\newblock In \emph{Proceedings of the IEEE/CVF Conference on Computer Vision
  and Pattern Recognition}, pages 13726--13735, 2020.

\bibitem[Xiao et~al.(2015)Xiao, Xia, Yang, Huang, and Wang]{Xiao_2015_CVPR}
Tong Xiao, Tian Xia, Yi~Yang, Chang Huang, and Xiaogang Wang.
\newblock Learning from massive noisy labeled data for image classification.
\newblock In \emph{Proceedings of the IEEE Conference on Computer Vision and
  Pattern Recognition (CVPR)}, 2015.

\bibitem[Xie et~al.(2016)Xie, Wang, Wei, Wang, and Tian]{xie2016disturblabel}
Lingxi Xie, Jingdong Wang, Zhen Wei, Meng Wang, and Qi~Tian.
\newblock Disturblabel: Regularizing {CNN} on the {Loss} {Layer}.
\newblock In \emph{Proceedings of the IEEE Conference on Computer Vision and
  Pattern Recognition}, pages 4753--4762, 2016.

\bibitem[Yun et~al.(2021)Yun, Oh, Heo, Han, Choe, and Chun]{Yun_2021_CVPR}
Sangdoo Yun, Seong~Joon Oh, Byeongho Heo, Dongyoon Han, Junsuk Choe, and
  Sanghyuk Chun.
\newblock Re-labeling imagenet: From single to multi-labels, from global to
  localized labels.
\newblock In \emph{Proceedings of the IEEE/CVF Conference on Computer Vision
  and Pattern Recognition (CVPR)}, pages 2340--2350, June 2021.

\bibitem[Zhang et~al.(2017)Zhang, Bengio, Hardt, Recht, and
  Vinyals]{zhang2019understanding}
C~Zhang, S~Bengio, M~Hardt, B~Recht, and O~Vinyals.
\newblock Understanding deep learning requires rethinking generalization.
\newblock In \emph{Proceedings of the International Conference on Learning
  Representation}, 2017.

\bibitem[Zoph et~al.(2018)Zoph, Vasudevan, Shlens, and Le]{zoph2018learning}
Barret Zoph, Vijay Vasudevan, Jonathon Shlens, and Quoc~V Le.
\newblock Learning transferable architectures for scalable image recognition.
\newblock In \emph{Proceedings of the IEEE Conference on Computer Vision and
  Pattern Recognition}, pages 8697--8710, 2018.

\end{thebibliography}
\end{document}